\documentclass[pdfusetitle,runningheads]{llncs}
\usepackage[T1]{fontenc}
\usepackage{graphicx}
\usepackage{booktabs} 
\usepackage{color}
\usepackage{amsmath}
\usepackage{nicefrac}
\usepackage{amssymb}
\usepackage[
  pdfusetitle
]{hyperref}

\begin{document}

\title{ICDAR 2026 HIPE-OCRepair Competition on LLM-Assisted OCR Post-Correction for Historical Documents}
\titlerunning{HIPE-OCRepair 2026}
%
\author{Maud Ehrmann\inst{1}\orcidID{0000-0001-9900-2193} 
\and
Emanuela Boros\inst{1}\orcidID{0000-0001-6299-9452} 
\and
Juri Opitz\inst{2}\orcidID{0000-0001-6892-4574}
\and
Andrianos Michail\inst{2}\orcidID{0009-0004-1025-7851}
\and
Florian Wagner\inst{2}
\and
Simon Clematide\inst{2}\orcidID{0000-0003-1365-0662}
}
\authorrunning{Ehrmann et al.}

\institute{
    École Polytechnique Fédérale de Lausanne (EPFL), Switzerland\\
    \email{}
    \and
    University of Zurich, Switzerland\\
    \email{}
}
\maketitle              
\begin{abstract}
We present the results of HIPE-OCRepair-2026, an ICDAR competition on LLM-assisted OCR post-correction of historical documents. OCR post-correction remains a long-standing challenge in digital heritage: large-scale collections of digitized documents are affected by legacy OCR errors, while re-digitization at scale remains impractical. Large language models (LLMs) offers a major opportunity to revisit this challenge, yet their effectiveness across languages, document types, and noise conditions --- and their tendency to hallucinate --- remains insufficiently understood.

HIPE-OCRepair-2026 pursues two objectives: (i) to evaluate the capabilities of modern OCR post-correction systems, and (ii) to provide a reproducible evaluation framework anchored in the HIPE-OCRepair-2026 dataset, a harmonized multilingual resource consolidating existing and newly curated historical datasets. Participants were tasked with correcting noisy OCR transcripts from historical newspapers and printed works in English, French, and German (17th--20th century), working at the level of coherent transcription units (paragraphs or articles) without access to source images. The evaluation adopts a retrieval-oriented rather than diplomatic scoring approach, reflecting the practical use case of search and access over digitized collections.

Four teams submitted systems ranging from zero-shot prompting to continued pre-training and fine-tuning, offering insights into the merits of different adaptation strategies. Results show that modern LLM-assisted systems can significantly improve OCR quality, but performance varies across datasets, languages, and noise levels. Overcorrection on low-noise inputs emerges as a recurring challenge, highlighting the importance of evaluation beyond character error reduction. The dataset, scorer, and evaluation pipeline are publicly released to support future research.

\end{abstract}

\section{Introduction}
\label{sec:introduction}

The large-scale digitization of historical collections has made millions of newspaper pages, books, and archival records searchable and available to researchers worldwide~\cite{neudecker_making_2016,beals_atlas_2020}. Yet the quality of the resulting text is uneven. Many collections were processed years or decades ago with OCR systems whose performance was limited by the technology of the time, and even modern engines continue to struggle with the specific challenges posed by historical documents: degraded paper and ink, non-standard typography, complex layouts, and multilingual content all contribute to recognition failures that remain difficult to avoid~\cite{jarlbrink_cultural_2017a,smith_research_2023}. 
The result is a systematic gap between what digitization has made available and what downstream applications --- from keyword search and information retrieval to named entity recognition and large-scale corpus analysis --- require to function reliably~\cite{chiron_impact_2017,vanstrien2020assessing,hamdi2020assessing,boros2022assessinga,ehrmann_named_2023}.

OCR post-correction, the automatic improvement of existing transcripts without reprocessing source images, is a long-standing response to this challenge~\cite{nguyen_survey_2021}. 
Despite decades of research spanning rule-based methods, statistical models, and neural sequence-to-sequence approaches~\cite{amrhein_supervised_2018,schaefer_twostep_2020}, the problem remains far from solved. 
Performance is highly sensitive to language, historical period, document type and noise characteristics, and robust generalization across heterogeneous collections has remained elusive~\cite{strobel_transformerbased_2022}. At the same time, the scale of the accumulated OCR debt --- the vast backlog of already-digitized pages with insufficient transcript quality --- makes re-digitization impractical, as reprocessing millions of pages requires infrastructure, funding, and coordination that most institutions cannot sustain.

The rise of large language models (LLMs) offers a new opportunity to revisit this challenge. Trained on large amounts of well-formed text, these models can perform noisy-to-clean transformations in zero-shot and few-shot settings. Their ability to exploit broad contextual information makes them promising candidates for correcting degraded historical transcripts at scale. Recent studies explore a range of approaches: prompt-based zero-shot correction with large proprietary and open models~\cite{gupta2021unsupervised,zhang_postocr_2024,boros_postcorrection_2024,kanerva_ocr_2025}, fine-tuning and instruction-tuning of encoder-decoder or decoder-only architectures on aligned OCR/ground-truth pairs~\cite{soper_bart_2021,thomas_leveraging_2024,kim2025early}, and hybrid pipelines combining error detection with generative correction~\cite{beshirov_postocr_2025,kim2025early}. Across these studies, fine-tuned models generally outperform zero-shot approaches, and metadata and contextual cues offer varying benefits. The broader picture, however, remains unclear: LLM performance varies considerably across languages, historical periods, and error types, and existing results are scattered across heterogeneous experimental settings that are difficult to compare directly~\cite{boros_postcorrection_2024,kanerva_ocr_2025}.

Crucially, OCR post-correction is not a form of unconstrained text generation, as historical documents are primary sources whose textual content should be preserved. A model that silently modernizes vocabulary or introduces plausible but absent content --- hallucinating rather than correcting --- is not improving a transcript but altering the source. Whether LLMs can reliably correct historical OCR across diverse collections without degrading or hallucinating content therefore remains an open question
A primary application scenario is information retrieval: users searching digitized archives depend on accurate transcriptions to locate relevant documents, and character-level errors, even at modest rates, can render words unsearchable and lead to incomplete retrieval results.

HIPE-OCRepair-2026\footnote{\url{https://hipe-eval.github.io/HIPE-OCRepair-2026/}} addresses these questions directly: Can modern LLM-based systems reduce OCR errors in legacy digitized collections while avoiding overcorrection and hallucination? Our evaluation campaign adopts a retrieval-oriented perspective, favoring linguistic accuracy over fully diplomatic transcription. In particular, it emphasizes the recovery of correct word forms over the faithful reproduction of typographic or layout features that are less consequential for search and text-based analysis. This design choice is reflected in the normalization applied prior to scoring (Section~\ref{sec:evaluation}).

OCR post-correction has previously been addressed in ICDAR shared tasks. 
The 2017 competition~\cite{chiron_icdar2017_2017} introduced the first systematic evaluation on historical newspaper data, and the 2019 edition~\cite{rigaud_icdar_2019a} expanded scope and refined the evaluation protocol. 
HIPE-OCRepair-2026 builds on this lineage while extending it in three ways. First, it explicitly targets LLM-based and other modern generative approaches to OCR correction. Second, it is built on HIPE-OCRepair-2026, a harmonized multilingual benchmark consolidating existing and newly curated datasets under unified segmentation and transcription guidelines, including selective correction of reference transcriptions where needed. Third, it provides a fully reproducible evaluation framework with standardized metrics, an official scorer, and an evaluation pipeline for systematic cross-system comparison.

The remainder of this paper is structured as follows. 
Section~\ref{sec:task} defines the task. 
Section~\ref{sec:data} describes HIPE-OCRepair-2026 dataset. Section~\ref{sec:evaluation} presents the evaluation framework. 
Section~\ref{sec:systems} describes the baseline and participating systems. 
Section~\ref{sec:results} presents and discusses the results. 
Section~\ref{sec:conclusion} concludes with findings and perspectives.

\section{Task Definition}
\label{sec:task}

HIPE-OCRepair-2026 is a post-correction task: given a noisy OCR transcript of a historical document, systems must produce a corrected transcription that more closely matches the manually verified reference. In line with standard practice in OCR post-correction, systems operate on text alone and have no access to the source document image. This reflects the practical reality of OCR debt scenarios, where the goal is to improve existing transcripts at scale without reprocessing original scans.

The input consists of an \emph{OCR hypothesis} and associated \emph{document metadata}. The OCR hypothesis is the raw transcription of a \emph{transcription unit} --- a coherent piece of text providing sufficient context for correction, typically  a paragraph or article --- which may contain character substitutions, insertions, deletions, and layout artefacts. 
The metadata specifies the language of the source document, as well as additional contextual information where available, such as the approximate date and publication title. 
The expected output is a corrected transcription of the OCR hypothesis. Systems are evaluated against a manually verified \emph{reference transcription} of the same transcription unit\footnote{See also: \url{https://github.com/hipe-eval/HIPE-OCRepair-2026-data/blob/main/README-Participation-Guidelines.md}}. 

This task presents particular challenges for LLM-based approaches. The input may contain severe distortions that obscure lexical and syntactic structure, yet systems must infer corrections from textual evidence alone, without access to the source image.  
More fundamentally, post-correction is a constrained task: systems are expected to restore intended word forms and improve transcription fidelity, but must not introduce content absent from the source or resolve ambiguities speculatively. This distinguishes the task sharply from free text generation or paraphrasing. It also means that characteristic properties of LLMs --- their fluency, their bias toward contemporary language, their capacity for hallucination --- can become potential liabilities as much as assets. These difficulties are compounded by the diversity of languages, historical periods, document types, and noise profiles across which systems must operate; concrete examples of OCR hypotheses and their reference transcriptions are available on the task website.


\section{Data: HIPE-OCRepair-2026 Dataset}
\label{sec:data}

The competition is grounded in the HIPE-OCRepair-2026 dataset, a harmonized multilingual benchmark of parallel OCR and ground-truth text pairs for historical documents. 
It covers three languages --- English, French, and German -- and two document types --- historical newspapers and printed books --- spanning the 17th to the 20th century. 
The dataset combines four substantially re-curated existing datasets with one dataset newly created for this shared task (\texttt{impresso-snippets}). Table~\ref{tab:datasets} provides an overview\footnote{See also \url{https://github.com/hipe-eval/HIPE-OCRepair-2026-data/blob/main/documentation/README.md}}.

\subsection{Curation Principles}

Existing datasets exhibit substantial variation in transcription policies, segmentation strategies, and ground-truth quality. All datasets were therefore processed through a unified curation pipeline comprising format standardization, quality filtering (removal of documents with CER $> 0.15$ or very short units), transcription-unit harmonization, and systematic manual revision of the development and test ground truths. 
Harmonization ensured that correction targets were neither too short (individual lines were avoided) nor excessively long aggregations of unrelated content. By contrast, the training sets retain their original transcriptions, with harmonization limited to segmentation and formatting.

A central curation decision was to adopt a semi-diplomatic transcription standard that prioritizes linguistically interpretable word forms while not preserving all historical graphemic or typographic detail, such as ligatures or non-modern letterforms like long \emph{s}.
In addition, whitespace artefacts inherited from legacy handling of hyphenated line breaks were regularized, for example where words split across lines had been represented with internal whitespace in the original ground truth. 
This choice reflects the primary application scenario of information retrieval, for which mildly standardized word forms are more useful than strict preservation of historical or typographic variation. 

Where layout information was present in the original data, it was retained in the benchmark files to support potential layout-aware use. In particular, soft hyphens were explicitly curated to mark word splits across line breaks, rather than being discarded as incidental layout artefacts. Systems were not required to reproduce layout information in their outputs. Full documentation of the curation process and transcription guidelines will be provided in a dedicated benchmark paper.

\subsection{Dataset Descriptions}

\subsubsection{\textbf{\texttt{dta19}}}
This dataset consists of pages sampled from 39 German books (18th--19th century) from the \textit{Deutsches Textarchiv} (DTA)~\cite{springmann_ground_2018a}, printed in Fraktur. 
Since the original dataset contains image/ground-truth pairs but no OCR hypothesis, OCR was generated by applying a Tesseract Fraktur model\footnote{\url{https://ub-backup.bib.uni-mannheim.de/~stweil/tesstrain/german_print_20231218/tessdata_best/german_print_20.traineddata}}~\cite{Weil:2021}. 
Because OCR on the original images was too accurate for a challenging post-correction setting, controlled image degradation was introduced at three noise levels: level~0 (no added noise), level~1 (targeting $\approx 3\%$ CER), and level~2 (targeting $\approx 7\%$ CER). 
Two sampling strategies were defined: \emph{matched} (same pages at all noise levels) and \emph{unmatched} (different pages at each level to avoid information leakage in the shared task). 
The official competition evaluation used the unmatched strategy.


\begin{table}[t]
\centering
\caption{Overview of the \textbf{HIPE-OCRepair-2026 dataset}.}
\label{tab:datasets}
\begin{tabular}{llllll}
\toprule
\textbf{Dataset part} & \textbf{Type} & \textbf{Lang} & \textbf{Period}  & \textbf{Origin} & \textbf{License} \\
\midrule
\texttt{dta19}             & book      & de        & 18C--19C   & existing  & CC BY-SA 4.0    \\
\texttt{icdar2017}         & newspaper & en, fr    & 17C--20C   & existing  & CC BY-NC-SA 4.0 \\
\texttt{impresso-snippets} & newspaper & de, en, fr & 19C--20C  & new       & CC BY-NC-SA 4.0 \\
\texttt{impresso-nzz}      & newspaper & de        & 18C--20C   & existing  & CC BY-NC 4.0    \\
\texttt{overproof}         & newspaper & en        & 19C--20C   & existing  & research only   \\
\bottomrule
\end{tabular}
\end{table}

\subsubsection{\textbf{\texttt{icdar2017}}}
This dataset consists of historical newspapers from the ICDAR 2017 OCR Post-Correction Competition~\cite{chiron_icdar2017_2017}, drawn from the National Library of France and the British Library, and digitized as part of the IMPACT project~\cite{balk_impact_2009}. The original material consists of very long concatenated text sequences and therefore required semantic chunking; low-quality chunks were filtered out (CER $> 0.15$). The ground truth of the development and test sets were manually re-corrected. Splits were derived from the original ICDAR 2017 train and test partitions.

\subsubsection{\textbf{\texttt{impresso-snippets}}} 
Newly created for this benchmark, this dataset consists of paragraph-level newspaper snippets (9--15 lines each) sampled from digitized collections in the Impresso project\footnote{\url{https://impresso-project.ch}}, covering German, French, and English materials from 1800--1959. Ground truth was produced through manual correction by trained annotators with access to paragraph images. Line information is preserved in full. New train, dev and test splits were created per language. For a few items, lines were reocrized using the same Tesseract model as for \texttt{dta19}.

\subsubsection{\textbf{\texttt{impresso-nzz}}} This dataset consists of front pages of the Neue Zürcher Zeitung (NZZ), spanning 1780--1947, published in German Fraktur (blackletter) type~\cite{strobel_improving_2019,strobel_ground_2019}. OCR was produced with ABBYY FineReader Server~11, and the ground truth was manually corrected. Pages with structural misalignments between OCR and ground truth were excluded. Layout information (line breaks, paragraph breaks, soft hyphens) is fully preserved. \texttt{impresso-nzz} is available as training data, but it was not part of the official shared task ranking because the underlying test data was publicly available prior to the competition.

\subsubsection{\textbf{\texttt{overproof}}} This dataset consists of English newspaper articles from the \textit{Sydney Morning Herald }(Trove / National Library of Australia) and \textit{Chronicling America} (Library of Congress), covering the 19th--20th century~\cite{evershed_correcting_2014a}. The ground truth is based on crowd-sourced Trove corrections, entirely manually verified and enriched with explicit soft hyphen annotations from our side, using the facsimile. As no original splits were available, train, development, and test splits were created for the HIPE-OCRepair-2026 benchmark. However, \texttt{overproof} is not included in the official competition ranking because the underlying test data was publicly available prior to the competition, and is made available as training data only.

\subsection{Dataset Statistics and Format}

Table~\ref{tab:splits} summarizes dataset composition, including split sizes, transcription-unit types, token counts, and average CER per test split. 
The competition test sets comprise 740 transcription units across eight test files, while the benchmark, including training and development splits, contains 995k tokens. CER values vary substantially across datasets and languages, ranging from near-zero noise in \texttt{dta19-l0} to moderate noise levels around 0.09 in \texttt{overproof}, reflecting the diversity of OCR conditions represented in the benchmark. 


\begin{table}[t]
\centering
\caption{Statistical profile of the HIPE-OCRepair-2026 dataset.
Tokens are reported across all splits; avg.\ CER (macro) is
reported for the test splits, after text normalization;
\textsuperscript{*}not part of the official ranking.}
\label{tab:splits}
\begin{tabular}{lllrrrrl}
\hline
\textbf{Dataset} & \textbf{Lang} & \textbf{Unit} &
    \textbf{Train} & \textbf{Dev} & \textbf{Test} &
    \textbf{Tokens} & \textbf{CER} \\
\hline
\texttt{dta19-l0} & de & page & 190 & 110 & 30 & 70,634 & 0.004 \\
\texttt{dta19-l1} & de & page & 190 & 110 & 30 & 73,231 & 0.024 \\
\texttt{dta19-l2} & de & page & 190 & 110 & 30 & 72,298 & 0.057 \\
\texttt{icdar2017}         & en & chunk     & 455 & 188 & 100 & 256,404 & 0.030 \\
\texttt{icdar2017}         & fr & chunk     & 391 &  -- & 100 & 198,881 & 0.018 \\
\texttt{impresso-snippets} & de & paragraph &  50 &  10 & 100 &  12,764 & 0.032 \\
\texttt{impresso-snippets} & en & paragraph &  50 &  10 & 100 &  15,253 & 0.017 \\
\texttt{impresso-snippets} & fr & paragraph &  50 &  10 & 100 &  13,013 & 0.018 \\
\hline
\texttt{impresso-nzz}\textsuperscript{*}  & de & page      & 150 &  -- &  17 & 221,103 & 0.099 \\
\texttt{overproof}\textsuperscript{*} & en & article   & 146 &  30 &  32 &  61,437 & 0.090 \\
\hline
\end{tabular}
\end{table}

All datasets are serialized in JSON Lines format, following a common schema,\footnote{\url{https://github.com/hipe-eval/HIPE-OCRepair-2026-data/blob/main/schema/hipe-ocrepair.schema.json}} with one JSON object per transcription unit. 
Each object contains \textit{document metadata} (identifier, language, date, publication title) and an \textit{OCR hypothesis} (text to correct, with optional sub-segmentation offsets at line, sentence, and chunk levels), together with either ground truth in reference files or post-correction output in system submission files. 


HIPE-OCRepair-2026 dataset is released on GitHub under version \texttt{v0.9}, with release tag \href{https://github.com/hipe-eval/HIPE-OCRepair-2026-data/releases/tag/v0.9.5}{\texttt{v0.9.5}}. Full statistics, including character counts and CER distributions per split, are available in the data repository.\footnote{\url{https://github.com/hipe-eval/HIPE-OCRepair-2026-data}}

\section{Evaluation Framework}
\label{sec:evaluation}

\subsection{Metrics}

\subsubsection{Primary metric: Character Match Error rate.}

Our primary evaluation metric is character-level Match Error Rate (cMER), defined as:
\[
\mathrm{cMER} = \frac{S + D + I}{H + S + D + I}
\]
where $H$ denotes hits, $S$ substitutions, $D$ deletions, and $I$ insertions in the character-level alignment between hypothesis and reference. Unlike standard Character Error Rate (CER), cMER is bounded in $[0,1]$ because insertions appear in the denominator, reducing sensitivity to over-generation~\cite{neudecker_ocrd_2019}. Character-level metrics are generally preferred over word-level metrics for historical OCR correction, as they better accommodate historical spelling variation.

cMER is reported at two levels of aggregation. The micro-averaged cMER (\texttt{cmer\_micro}) sums counts ($H$, $S$, $D$, $I$) across all transcription units in a dataset before computing the ratio, thereby giving proportionally more weight to longer units.
\texttt{cmer\_micro} is the primary system ranking criterion. 
We additionally report the macro-averaged cMER (\texttt{cmer\_macro}), the arithmetic mean of per-unit scores, which gives equal weight to each transcription unit regardless of length.

\subsubsection{Secondary metric: preference score}
Aggregate error rates such as cMER summarize the overall \emph{magnitude} of error reduction but can mask inconsistent behavior: a system that corrects a few documents substantially while degrading many others may still achieve a favorable average score. 
To capture the \emph{consistency} of improvement across transcription units, we complement cMER with a sign-based item-level preference score.

For each transcription unit $i$, the score is defined as:
\[
s_i = \operatorname{sign}\!\big(\mathrm{cMER}_{\mathrm{ocr},i}
      - \mathrm{cMER}_{\mathrm{cor},i}\big) \in \{+1, 0, -1\},
\]
where $\mathrm{cMER}_{\mathrm{ocr},i}$ is the error rate of the OCR hypothesis and $\mathrm{cMER}_{\mathrm{cor},i}$ is that of the post-correction output. 
A score of $+1$ indicates improvement over the OCR hypothesis, $0$ no change, and $-1$ degradation. 
The reported metric (\texttt{pref\_score\_cmer\_macro}) is the macro-average of $s_i$ across all transcription units, so each unit contributes equally and large gains on a few units cannot dominate the score. 
The metric therefore complements cMER by highlighting how often post-correction helps or harms individual transcription units.

\subsubsection{Additional metrics and confidence intervals}

We also report word-level MER at both micro and macro aggregation levels (\texttt{wmer\_micro}, \texttt{wmer\_macro}), computed over word-level alignments under the same normalization used for character-level scoring. All reported metrics are accompanied by 95\% bootstrap confidence intervals based on 10{,}000 resamples of transcription units, enabling assessment of sampling uncertainty and supporting assessment of result stability.

\subsection{Text Normalization Before Scoring}

Before scoring, both the post-correction output and the reference transcription are normalized in two steps. 
First, \emph{layout normalization} removes soft hyphens followed by line breaks ($\neg\backslash$n) and joins the surrounding word parts; remaining line breaks 
are converted to spaces.
Second, \emph{IR-style normalization} lowercases the text, replaces punctuation and other non-word characters with spaces, and collapses repeated whitespace to a single space. 
The evaluation is thus case- and punctuation-insensitive, but remains sensitive to accented characters (e.g.\ \textit{é} and \textit{e} are distinct). 
This normalization policy reflects the IR application scenario of the benchmark where indexing systems often apply comparable transformations. 
Systems are therefore not penalized for layout differences in their output.

\subsection{Ranking Protocol And Submission Rules}

\subsubsection{Per-dataset and overall rankings}

Scoring is performed per dataset, using \texttt{cmer\_micro} as the primary criterion and \texttt{pref\_score\_cmer\_macro} as the secondary criterion. 
The official competition ranking is computed as a weighted mean of per-dataset \texttt{cmer\_micro} scores across the eight official test sets, with weights chosen to balance language-level contributions. English and French each contribute two equally weighted test sets. For German, \textit{impresso-snippets} carries weight~1, while the three \texttt{dta19} noise levels (\texttt{dta19-l0}, \texttt{dta19-l1}, \texttt{dta19-l2}) carry weight~$\nicefrac{1}{3}$ each, so that the three \texttt{dta19} sets together contribute the same total weight as one other test set. Per-language rankings are reported using the same weighting logic within each language.

\subsubsection{Submission rules}

Teams were allowed to submit up to three runs per dataset and language. External resources and pre-trained models were permitted provided they were documented. If a system did not provide a post-correction output for a transcription unit, the original OCR hypothesis was used unchanged for scoring.

\subsection{Evaluation Infrastructure}

The evaluation toolkit consists of two components. 
The \emph{HIPE-OCRepair scorer}\footnote{\url{https://github.com/hipe-eval/HIPE-OCRepair-scorer}} is a Python package\footnote{\url{https://pypi.org/project/hipe-ocrepair-scorer/}} that computes all reported metrics, enabling participants to evaluate their systems locally before submission. 
The \emph{evaluation repository}\footnote{\url{https://github.com/hipe-eval/HIPE-OCRepair-2026-eval}} contains the official test data (also present in the data repository), all system submissions, scoring scripts, and a Makefile for reproducing the full evaluation pipeline, ensuring transparent and reproducible scoring.

\section{System Descriptions}
\label{sec:systems}

\begin{table}[t]
\centering
\caption{Overview of participating systems.}
\label{tab:systems}
\begin{tabular}{llrc}
\hline
\textbf{Team} & \textbf{Model} & \textbf{Param.} & \textbf{Runs} \\
\hline
\textsc{BLOCR}          & DeepSeek / Gemma~3    & --  &  8 \\
\textsc{BnF-Mistral}    & Mistral Small~3       & 24B & 24 \\
\textsc{Zakaria-ENSIAS} & Qwen3                 & 8B  &  7 \\
\textsc{L3i}            & Qwen3                 & 14B & 11 \\
\hline
\end{tabular}
\end{table}

\subsection{Baseline}

As a minimal reference, we include a no-correction baseline that returns the OCR hypothesis unchanged. 
Its performance provides a lower bound for system comparison and directly reflects the underlying OCR quality of each test set, as measured by the initial cMER. It is particularly informative in low-noise settings, where unnecessary modifications by post-correction systems can degrade otherwise accurate transcriptions.

\subsection{Participating Systems}

Four teams submitted a total of 50 distinct runs, covering all official test datasets and languages. Table~\ref{tab:systems} summarizes the submitted systems. 
The approaches span a broad methodological spectrum: from zero-shot prompting of off-the-shelf models to continued pre-training followed by supervised fine-tuning and multi-step inference pipelines.  All teams used decoder-only large language models ranging from 8B to 24B parameters. 
A common thread across all submissions is the use of task-specific prompting designed to guide the model toward faithful correction rather than free rewriting; teams differ substantially, however, in how much they further adapt the model to the task and in the strategies they employ to control hallucination and output quality. The following system descriptions are compiled from information provided by the participants.\footnote{Participating teams may publish further technical details in separate system papers.}

\subsubsection{Zakaria-ENSIAS}

Team \textsc{Zakaria-ENSIAS}, affiliated with \textit{ENSIAS} (Morocco), submitted 7 runs covering all official test sets except \texttt{dta19-l2}. Their system performs zero-shot OCR post-correction using Qwen3-8B served locally via Ollama. Each document is processed independently: a prompt constructed from the OCR hypothesis and document metadata (language, publication date, document type) instructs the model to correct character recognition errors, fix hyphenation artefacts and spurious word splits, remove garbled fragments, and preserve historical spelling without modernization. A temperature of 0.1 favors conservative, faithful corrections. To guard against degenerate outputs, a length-ratio safeguard rejects responses outside a predefined range relative to the input length (accepting outputs between 0.35$\times$ and 2.8$\times$ the input length) and falls back to the original OCR text after up to two retries. A post-processing step strips chain-of-thought content and any echoed prompt fragments from the model output. No fine-tuning, retrieval augmentation, or external lexicons were used.

\begin{table}[t]
\centering
\caption{Methodological features of participating systems.}
\label{tab:approaches}
\begin{tabular}{lcccc}
\hline
\textbf{Feature} & \textsc{Zakaria} & \textsc{L3i} &
    \textsc{BLOCR} & \textsc{BnF-Mistral} \\
\hline
Approach & zero-shot & fine-tuned & zero-shot & CPT + FT \\
Training data  & competition  & competition & selection & comp.\ + synth. \\
\hline
Continued pre-training  & --         & --         & --         & \checkmark \\
Fine-tuning             & --         & \checkmark & --         & \checkmark \\
Synthetic data          & --         & --         & --         & \checkmark \\
Task-specific prompt    & \checkmark & \checkmark & \checkmark & \checkmark \\
Metadata in prompt      & \checkmark & --          & --         & --         \\
Error type guidance     & \checkmark & \checkmark & \checkmark & --         \\
Output validation       & \checkmark & --         & \checkmark & \checkmark \\
Hallucination control   & \checkmark & partial    & --         & \checkmark \\
Routing to larger model & --         & --         & --         & \checkmark \\
\hline
\end{tabular}
\end{table}

\subsubsection{L3i}

Team \textsc{L3i}, affiliated with the \textit{University of La Rochelle} (France), submitted 11 runs: run~1 covered all official test sets, while run~2 was submitted for the \texttt{impresso-snippets} datasets only (all three languages). Both runs fine-tuned Qwen3-14B with instruction tuning on all available training pairs, using a single prompt written in English across all languages and datasets. The prompt provided the OCR text to correct, with instructions to preserve meaning, historical spelling, names, numbers, and dates, and to avoid paraphrasing. No document metadata was included. Long inputs were split into non-overlapping chunks of 256 tokens, processed independently during both training and inference.

Run~2 explored a detection-augmented variant in which likely erroneous tokens were first identified by a token classification model and inserted into the prompt as explicit correction targets, with all other settings identical to run~1. It underperformed run~1, which the team attributed to insufficient training data for the more complex pipeline.

\subsubsection{BLOCR}

Team \textsc{BLOCR}, affiliated with the \textit{British Library} (United Kingdom), submitted 8 runs, corresponding to a single approach applied across all official test sets. Their approach begins with a model selection phase: six candidate models were evaluated via the Hugging Face Inference API for inference stability and performance, using a sample of approximately 10k tokens per language drawn from the training data. DeepSeek was selected for English and Gemma~3 for French and German based on cMER and preference score on this sample.

The same prompt was used for both model selection and final submission. It combines the OCR hypothesis with the document language and instructions to correct four error classes: over-segmentation, under-segmentation, misrecognized characters, and missing characters, following the taxonomy of~\cite{soper_bart_2021}. Document metadata such as date, publication title, and document type were not included. The model was instructed to return a corrected version of the text alongside a CSV file listing detected errors, classified by type and confidence level; the latter was produced for analysis purposes and did not feed back into the correction pipeline. Generation used a temperature of 0.2 with no other parameters explicitly set beyond a token buffer to avoid truncation.

\subsubsection{BnF-Mistral}

Team \textsc{BnF-Mistral}, affiliated with the \textit{National Library of France} (France) and the \textit{Mistral AI}, submitted 24 runs with the most elaborate pipeline of the competition. Their base model is a version of Mistral Small~3 (24B) that underwent continued pre-training on historical BnF documents predating 1900 (ca 80B tokens), and broader multilingual corpora. An intermediate post-training checkpoint was then fine-tuned using LoRA on approximately 10,000 samples drawn from the competition datasets, supplemented by synthetic OCR pairs in French, English, and German. The latter were generated by artificially degrading periodical documents from the 18th to 20th century using confusion matrices and insertion/deletion probabilities derived from the competition data, retaining only hard examples after filtering. Different runs reflect different data mixtures and LoRA configurations. 

At inference, language-specific system prompts are applied, with more detailed anti-hallucination constraints for German and English. An iterative judge-and-retry loop flags suspicious corrections (hallucination, missing text, overcorrection) and triggers up to three retries, reducing the flagging rate from 42\% to 6\%. Documents where the CER between raw OCR and corrected output exceeds 15\% are additionally routed to a larger model. The participating team reports that the full post-processing pipeline yields a 27\% improvement in cMER on the validation splits.

\begin{table}[tb]
\centering
\caption{Overall competition ranking on the test split. For readability, only distinct runs are shown; runs whose predictions were identical to another run for most official test sets are omitted from this summary table. Lower cMER is better; higher preference score is better. Full results, including 95\% confidence intervals and omitted runs, are available in the online evaluation report.}
\label{tab:overall-results}
\begin{tabular}{lccc}
\toprule
\textbf{Rank} & \textbf{System} & \textbf{cMER} $\downarrow$ & \textbf{Pref} $\uparrow$ \\
\midrule
1 & \textsc{BnF-Mistral} (run1) & 0.0050 & 0.9000 \\
2 & \textsc{BnF-Mistral} (run3) & 0.0071 & 0.8737 \\
3 & \textsc{BnF-Mistral} (run2) & 0.0090 & 0.8739 \\
4 & \textsc{BLOCR} (run1)       & 0.0106 & 0.7028 \\
5 & \textsc{L3i} (run1)         & 0.0176 & 0.3591 \\
6 & baseline           & 0.0226 & 0.0000 \\
\bottomrule
\end{tabular}
\end{table}

\subsection{Summary of Approaches}

Table~\ref{tab:approaches} maps the main methodological features across teams. The approaches span three broad levels of task-specific adaptation: continued pre-training combined with fine-tuning and a multi-step inference pipeline (\textsc{BnF-Mistral}); supervised fine-tuning with task-specific prompting (\textsc{L3i}); and zero-shot prompting with output validation strategies (\texttt{Zakaria-ENSIAS}, \textsc{BLOCR}). Among all submissions, \textsc{BnF-Mistral} achieves the strongest overall performance, consistent with the depth of its adaptation to the task. The two zero-shot systems invest in complementary strategies --- metadata use, error type guidance, retry loops, and output length constraints --- that partially compensate for the absence of task-specific training. The impact of individual design choices, including the use of document metadata, explicit error modeling, and hallucination control mechanisms, is examined in detail in the next Section on results.

\section{Results and Discussion}
\label{sec:results}

As shown in Table~\ref{tab:overall-results}, the overall ranking is dominated by variants of \textsc{BnF-Mistral}. 
The no-correction baseline ranks last, which is notable because abstaining from any edits is a competitive strategy in low-noise settings~\cite{boros_postcorrection_2024}.
Notably, the best \textsc{BnF-Mistral} runs achieve preference scores close to 0.9, indicating highly consistent improvements.

\begin{table}[tb]
\centering
\caption{Per-language ranking (test split). Scores are weighted means over the official test sets for each language. Only distinct runs with complete coverage for the respective language are shown.
}
\label{tab:per-language-results}
\begin{tabular}{llcc}
\toprule
\textbf{Lang} & \textbf{System} & \textbf{cMER} $\downarrow$ & \textbf{Pref} $\uparrow$ \\
\midrule
de & \textsc{BnF-Mistral} (run1)    & 0.0061 & 0.8300 \\
de & \textsc{BnF-Mistral} (run3)    & 0.0089 & 0.7411 \\
de & \textsc{BnF-Mistral} (run2)    & 0.0101 & 0.9017 \\
de & \textsc{BLOCR} (run1)          & 0.0153 & 0.5183 \\
de & \textsc{L3i} (run1)            & 0.0202 & 0.4322 \\
de & baseline              & 0.0286 & 0.0000 \\
\midrule
en & \textsc{BnF-Mistral} (run1)    & 0.0046 & 0.9400 \\
en & \textsc{BnF-Mistral} (run3)    & 0.0055 & 0.9300 \\
en & \textsc{BLOCR} (run1)          & 0.0070 & 0.8900 \\
en & \textsc{BnF-Mistral} (run2)    & 0.0083 & 0.8850 \\
en & \textsc{Zakaria-ENSIAS} (run1) & 0.0176 & 0.6250 \\
en & \textsc{L3i} (run1)            & 0.0180 & 0.3200 \\
en & baseline              & 0.0215 & 0.0000 \\
\midrule
fr & \textsc{BnF-Mistral} (run1)    & 0.0042 & 0.9300 \\
fr & \textsc{BnF-Mistral} (run3)    & 0.0068 & 0.9500 \\
fr & \textsc{BnF-Mistral} (run2)    & 0.0088 & 0.8350 \\
fr & \textsc{BLOCR} (run1)          & 0.0096 & 0.7000 \\
fr & \textsc{L3i} (run1)            & 0.0145 & 0.3250 \\
fr & baseline              & 0.0176 & 0.0000 \\
fr & \textsc{Zakaria-ENSIAS} (run1) & 0.0185 & 0.5950 \\
\bottomrule
\end{tabular}
\end{table}

Table~\ref{tab:per-language-results} shows that this pattern holds across all three languages. \textsc{BnF-Mistral} (run1) ranks first for German, English, and French, with cMER scores of 0.0061, 0.0046, and 0.0042, respectively. The strongest alternative system is \textsc{BLOCR}, which ranks fourth for German, third for English, and fourth for French. English shows the closest competition among the top systems, with \textsc{BLOCR} approaching the \textsc{BnF-Mistral} runs more closely than in the other language-level rankings.

\begin{table}[t]
\centering
\caption{Best-performing system(s) per dataset on the primary metric (\texttt{cmer\_micro}), together with the no-correction baseline. Where the top-ranked system was not significantly better than the next-ranked system under paired bootstrap testing, both systems are reported. $\Delta$ denotes the absolute improvement over the no-correction baseline.}
\label{tab:best-per-dataset}
\begin{tabular}{l l c c c}
\toprule
\textbf{Dataset} & \textbf{Best system(s)} & \textbf{cMER} & \textbf{Baseline} & \textbf{$\Delta$} \\
\midrule
\texttt{dta19-l0} (de) & \textsc{BnF-Mistral} (run2) & 0.0029 & 0.0040 & 0.0011 \\
\texttt{dta19-l1} (de) & \textsc{BnF-Mistral} (run1) & 0.0054 & 0.0240 & 0.0186 \\
\texttt{dta19-l2} (de) & \textsc{BnF-Mistral} (run1) & 0.0082 & 0.0546 & 0.0464 \\
\texttt{icdar2017} (en) & \textsc{BnF-Mistral} (run1) & 0.0044 & 0.0260 & 0.0216 \\
\texttt{icdar2017} (fr) & \textsc{BnF-Mistral} (run1) & 0.0040 & 0.0184 & 0.0144 \\
\texttt{impresso-snippets} (de) & \textsc{BnF-Mistral} (run1) & 0.0058 & 0.0296 & 0.0238 \\
\texttt{impresso-snippets} (en) & \textsc{BnF-Mistral} (run1), \textsc{BLOCR} (run1) & 0.0049 & 0.0170 & 0.0121 \\
\texttt{impresso-snippets} (fr) & \textsc{BnF-Mistral} (run1) & 0.0044 & 0.0169 & 0.0125 \\
\bottomrule
\end{tabular}
\end{table}

Table~\ref{tab:best-per-dataset} summarizes the best-performing system(s) for each official test set together with the no-correction baseline, making baseline-relative gains directly visible. The strongest systems reduce cMER on all datasets, but the absolute gain depends strongly on initial OCR quality, ranging from 0.0011 cMER on the already low-noise \texttt{dta19-l0} condition to 0.0464 on \texttt{dta19-l2}. A \textsc{BnF-Mistral} run is top-ranked on all eight official test sets. On seven of them, the top \textsc{BnF-Mistral} run is significantly better than the next-ranked system under paired bootstrap testing on \texttt{cmer\_micro}. The only exception is \texttt{impresso-snippets} English, where \textsc{BnF-Mistral} (run1) and \textsc{BLOCR} (run1) are statistically indistinguishable at the top, both reaching a rounded cMER of 0.0049.


Overall, the results show that LLM-based OCR post-correction can substantially improve retrieval-oriented text quality across heterogeneous historical collections. At the same time, the variation across datasets confirms the importance of reporting baseline-relative gains: a low absolute cMER may reflect either strong correction or an already accurate OCR hypothesis. The no-correction baseline is therefore not a trivial reference point, but an indicator of dataset difficulty and over-correction risk. The consistent outperformance of \textsc{BnF-Mistral} variants across all conditions reflects the depth of task-specific adaptation described in Section~\ref{sec:systems}: the gap between the top fine-tuned system and the best zero-shot alternative (\textsc{BLOCR}) is itself a finding about the relative merits of different adaptation strategies for this task.

\section{Conclusion}
\label{sec:conclusion}

HIPE-OCRepair-2026 set out to assess whether modern LLMs can effectively reduce OCR error in historical document collections under realistic constraints, and to provide a reproducible benchmark for this task.

The results provide a clear, though nuanced, answer. Across most datasets, LLM-based systems, in particular the \textsc{BnF-Mistral} runs, achieve substantial gains over the raw OCR input, represented by a no-correction baseline, demonstrating that LLMs can effectively reduce accumulated OCR debt without access to source images, even in multilingual and historically heterogeneous settings.

At the same time, performance is not uniform. Gains are strongest under moderate to high OCR noise, whereas low-noise historical text remains inherently sensitive to over-correction: when the input is already relatively accurate, even small unnecessary edits can degrade the transcription. Performance also varies across languages and datasets, with some settings (notably English and French datasets) showing tighter competition at the top and smaller margins between systems. The preference score further shows that successful post-correction is not only a matter of reducing average error, but also of avoiding degradation at the level of individual transcription units. 
This highlights a central tension in LLM-based approaches: while their generative capacity enables effective correction, it also introduces a risk of over-correction or unintended alteration, which must be controlled.

The design of the benchmark --- in particular its information-retrieval-oriented normalization --- shapes the interpretation of these results. By abstracting away from orthographic variation and layout features, the evaluation emphasizes lexical recoverability and downstream usability rather than strict diplomatic fidelity. Reported gains should therefore be understood primarily as improvements in searchability and text-based processing, rather than full restoration of historically faithful transcriptions.

Beyond the competition itself, the HIPE-OCRepair-2026 dataset, scorer, and evaluation toolkit constitute a set of durable, openly available assets for the community. These will be further consolidated and extended into a comprehensive benchmark, including a public leaderboard.

Future work should focus on improving robustness across languages and domains, developing methods to control overcorrection and hallucination, and extending evaluation to settings that require stricter fidelity to historical form. The integration of layout-aware and image-informed approaches also remains an important direction for bridging the gap between post-correction and full OCR reprocessing, especially given recent evidence that visually degraded inputs can also induce OCR hallucinations in vision-language models~\cite{he2025seeing}.



\begin{credits}
\subsubsection{\ackname} 
The authors thank the ICDAR 2026 Conference and Competition Committees for hosting the task and Corina Raclé for helping with the manual correction of the \texttt{impresso-snippets} data. This work was carried out within the \href{https://impresso-project.ch}{Impresso} project  (SNSF grant \href{https://data.snf.ch/grants/grant/213585}{No.\ CRSII5\_213585} and FNR grant No.\ 17498891)


\subsubsection{\discintname}
The authors have no competing interests to declare that are relevant to the content of this article. 
\end{credits}

%
%
%
\bibliographystyle{splncs04}
\bibliography{2026-HIPE-OCRepair-Benchmark}

\end{document}